\documentclass{sig-alternate}
\newfont{\mycrnotice}{ptmr8t at 7pt}
\newfont{\myconfname}{ptmri8t at 7pt}
\let\crnotice\mycrnotice%
\let\confname\myconfname%

\usepackage[german,american]{babel}
\usepackage[T1]{fontenc}
\usepackage[utf8]{inputenc}
\usepackage{times}
\usepackage{textcomp}
\usepackage{graphicx}
\usepackage{amssymb}
\usepackage{microtype} 

\usepackage{url}
\renewcommand{\tt}{\fontencoding{OT1}\fontfamily{cmtt}\selectfont}

\usepackage[usenames,dvipsnames]{color}
\usepackage{booktabs}
\usepackage[numbers,sectionbib]{natbib}
\makeatletter
\def\@copyrightspace{\relax}
\makeatother

\usepackage{footnote}

\begin{document}

\title{Machine Learning Based Detection of Clickbait Posts in Social Media}
\subtitle{
Submission for the Clickbait Challenge 2017}

\numberofauthors{4}
\author{
\alignauthor
Xinyue Cao\thanks{The work was performed when the author visited Penn State as a summer intern during the summer of 2017.}\\ \vspace{0.1in}
\affaddr{Beijing University of Posts and Telecommunications, China}\\ \vspace{0.05in}
\affaddr{xinyue.c@outlook.com}\\
\alignauthor
Thai Le\\ \vspace{0.1in}
\affaddr{Penn State University, USA}\\ \vspace{0.05in}
\affaddr{thai.le@ist.psu.edu}\\
\alignauthor
Jason (Jiasheng) Zhang\\ \vspace{0.1in}
\affaddr{Penn State University, USA}\\ \vspace{0.05in}
\affaddr{jpz5181@ist.psu.edu}\\
\and
\alignauthor
Dongwon Lee\\ \vspace{0.1in}
\affaddr{Penn State University, USA}\\ \vspace{0.05in}
\affaddr{dlee@ist.psu.edu}\\
}

\maketitle

\begin{abstract}
Clickbait (headlines) make use of misleading titles that hide critical information from
or exaggerate the content on 
the landing target pages 
to entice clicks. 
As clickbaits often use eye-catching wording to attract viewers, target contents are often of low quality.
Clickbaits are especially widespread on social media such as Twitter, adversely impacting user experience by causing immense dissatisfaction. Hence, it has become increasingly important to put forward a widely applicable approach to identify and detect clickbaits. In this paper, we make use of a dataset from the clickbait challenge 2017 (clickbait-challenge.com) comprising of over 21,000 headlines/titles, each of which is annotated by at least five judgments from crowdsourcing on how clickbait it is. We attempt to build an effective computational clickbait detection model on this dataset. We first considered a total of 331 features, filtered out many features to avoid overfitting and improve the running time of learning, and eventually  selected the 60 most important features for our final model. Using these features, Random Forest Regression achieved the following results: MSE=0.035 MSE, Accuracy=0.82, and F1-sore=0.61 on the clickbait class. 
\end{abstract}

\section{Introduction}

The media landscape is currently undergoing tremendous changes with news format moving from paper to online media. As such, online new headlines are being optimized in real-time, recasting  teaser messages to maximize click-through. Such online headlines are different from the traditional printed frontpage headlines, in which feedbacks contributing to the newspaper sales is often indirect, delayed, and incomplete \cite{Peter2012}. 
Hence, online news providers are trying to make headlines to be more attractive. In particular, some headlines are carefully worded to be eye-catching and often misleading, resulting in unsatisfactory user experience on social media. In addition, the content on landing pages (by yellow journalism) are of low quality and significantly under-deliver the content promised by the exaggerated headlines. 

We informally consider this type of online news headlines as {\em clickbait}. According to the Oxford English Dictionary, clickbait is defined as ``content whose main purpose is to attract attention and encourage visitors to click on a link to a particular web page." Clickbait usually leads to a site by withholding the promised ``bait." Typical example clickbaits  are as follows: ``{\tt What This Person Did For That Will AMAZE You!}” or ``{\tt 9 out of 10 Americans Are Completely Wrong About This Mind-Blowing Fact}."

This type of irresponsible reporting not only frustrates users, but also violates journalistic code of ethics. Scholars have argued that  current trend towards a merging of commercial and editorial interests is detrimental to democratic values~\cite{couldry2014advertising}. In this paper~\cite{couldry2014advertising}, Couldry and Turow offer a preliminary discussion of clickbait as an example of false or misleading news, and review the identifying characteristics and potential methods to detect clickbaits. Due to the importance of the problem and its implication, in recent years, much research has investigated on the detection of clickbaits.
For instance, Gianotto~\cite{Alison2014},  implements a browser plug-in that transcribes clickbait teaser messages based on a rule set so that they convey a more truthful, or rather ironic meaning. Beckman~\cite{Jake2015}, Mizrahi~\cite{Alex2015}, Stempeck~\cite{Matt2015}, Kempe~\cite{Rebecca2015} manually re-share clickbait teaser messages, adding spoilers. 

To address the problem of clickbait detection, 
capitalizing on informative patterns found in previous works, we propose a set of new features extracted from post text (e.g., tweets), target paragraphs, keywords of target paragraphs, similarity between post text and target content. 
Experimental results show that our model achieves 0.035 in MSE, 0.82 in Accuracy, and 0.61 in F1-score on clickbait class, respectively. 

Our main contribution is the extraction of novel features for clickbait classification which have not been previously studied, and show that these features are among the most effective indicators for  clickbait headlines. While still useful, however, much more research is needed toward the development of a detection method that is practical enough in real settings.

\section{Related Work}
There have been extensive number of studies on evaluation of the authenticity of news or articles on the web, especially fake news. But clickbait does not necessarily have false content. It usually appears as misleading and exaggerating headline or title on top of a genuine article. In fact, the concept of clickbait is originated from the advent of tabloid journalism focusing on sensationalization of soft news, which was claimed by Rowe~\cite{rowe2011obituary} on the properties of making changes on professional journalism.

In the field of psychology, information-gap theory was put forward to explain the curiosity arising as a gap in one’s knowledge, of which clickbait exploits to get more clicks. According to Loewenstein~\cite{loewenstein1994psychology}, riddles, events with unknown sequences, expectation violations or forgotten information are identified as stimuli that may spark involuntary curiosity. There have been also studies on analyzing structure of headlines which contain the properties such as sensationalism, luring, dramatization, emotionalism, etc..

Recently, there have been several literatures focusing on clickbait detection. Martin et al.~\cite{potthast2016clickbait} claimed to introduce the first machine learning approach to the clickbait detection problem, especially in the social network context. They were collecting the first clickbait corpus of 2992 tweets, comprising of 767 clickbait tweets annotated by three assessors. Markus~\cite{Markus} prevented links that relate to a fixed set of domains from appearing on the users’ news feeds. This rule-based approach is not scalable and may require continuous tunings accordingly with the emergence of new clickbait phrases. This might also block texts that are not necessarily clickbaits. Abhijnan et al.~\cite{chakraborty2016stop} achieved 93\% accuracy in detecting clickbaits. However, they do feature analysis on the whole dataset instead of a separated training set. Md Main Uddin Rony et al.~\cite{rony2017diving} only considered features extracted from the titles. However, our method considers the features from title, target, as well as the relation between them.

\section{Feature Engineering}
In this section, we share some of the novel features that we have extracted. In addition to these novel features, we also consider features reported in recent literature, summarized in Table.~\ref{table:totalFeatures}.

\begin{table}
\centering
\begin{tabular}{|p{4cm}|p{3cm}|}
\hline
Features & Sources\\ 
\hline
Readability (Flesch-Kincaid) on post and target & Boilerpipe et al.~\cite{kohlschutter2010boilerplate}\\
\hline
Basic statistics features, including total length in words on post text and target paragraphs, average length of word on post text, length of the longest word on post text. & Abjijnan et al.~\cite{chakraborty2016stop}, Martin et al.~\cite{potthast2016clickbait}, Prakhar et al.~\cite{biyani20168}\\
\hline
Whether start with number, on post text & Abjijnan et al.~\cite{chakraborty2016stop}, Martin et al.~\cite{potthast2016clickbait}, Prakhar et al.~\cite{biyani20168}\\ 
\hline
Whether start with 5W1H, on post text & Prakhar et al.~\cite{biyani20168}\\ 
\hline
Ratio of stop words, on post text & Abjijnan et al.~\cite{chakraborty2016stop}, Martin et al.~\cite{potthast2016clickbait}, Prakhar et al.~\cite{biyani20168}\\ 
\hline
Sentiment score (Stanford NLP), on post text & Martin et al.~\cite{potthast2016clickbait}\\ 
\hline
Internet slangs, on post text & Abjijnan et al.~\cite{chakraborty2016stop}, Martin et al.~\cite{potthast2016clickbait} \\ 
\hline
POS n-gram and POS pattern, on post text & Abjijnan et al.~\cite{chakraborty2016stop}\\ 
\hline
\end{tabular}
\caption{Features from previous papers} 
\label{table:totalFeatures}
\end{table}

\subsection{New Features}
\subsubsection{On relation between post text and target article content}
For clickbait, the target content might be of little value but the post text usually appears to be attractive. Hence, the differences between the target and post should be considered as a feature. In fact, we designed three features corresponding to the relation between the post and target as follows.

\begin{itemize}
\item \textbf{Similarity between post text and target title.} Considering the relation between post text and target article, the most direct approach is to compare the post text with the target title. We use the Perl module~\cite{PerlModule} that measures the similarity of two files or two strings based on the number of overlapping words, scaled by the lengths of the files. Text similarity is based on counting of overlapping words between the two files, and is normalized by the length of the files. The feature value is seen to be larger in case of non-clickbait than clickbait class. In fact, this feature is ranked 18 out of 180 in later features evaluation.

\item \textbf{Similarity between post text and target paragraphs.} With the same idea, another approach to consider the relation between post text and target article is to measure the similarity between post text and target paragraphs. Therefore, we make use of the same algorithm, the Perl module, to calculate such feature. The feature is ranked 47 out of 180 in later features evaluation. 

\item \textbf{Match between post text and target keywords.} The number of keywords of target paragraphs that are included in post text is calculated as the value of this feature. Keywords can represent key information of the target paragraphs to a very large degree. Noticeably, non-clickbait titles have many more matched keywords with target paragraphs, while clickbait ones have a relatively low matched phrases. The feature is ranked 37 out of 180 in later features evaluation.
\end{itemize}

\subsubsection{On post text}
Post text, which can be misleading or ambiguous in clickbaits, is the text posted linking a link to target content. Following are three features that have not been studied by previous literatures.

\begin{itemize}
\item \textbf{Existence of Part-Of-Speech (POS) pattern: NUMBER + NOUN PHRASE + VERB.} Information can be hidden as numeric forms. An instance displaying this feature would be: ``{\tt 10 things Apple will never tell you about iPhone}." Numbers in the post hide specific information, which eventually exploit the curiosity of the readers to make them commit clicking. The previous paper~\cite{Jake2015} considered a feature checking whether the title starts with a number. However, not every texts starting with a number manifests this malicious behavior. For instance, ``{\tt 2017 Chinese Horoscope Chicken Prediction}." Therefore, we extract this features from post text by using part-of-speech tagging with the pattern: Number\_NP\_VB . This feature is ranked 56 out of 180 in later features evaluation.

\item \textbf{Existence of POS pattern: NUMBER + NOUN PHRASE + THAT.}
Similarly, The above behavior can be manifested under the pattern: Number\_NP\_THAT. For example, ``{\tt 10 things that Apple will never tell you about iPhone}." In the latter experiment of This feature is ranked 77 out of 180 in later features evaluation.

\item \textbf{Number of tokens.} Many of previous literatures extract total number of word separated by whitespace in the post text as a feature, but here we tokenize post text by both whitespace and punctuation, which eventually also counts the number of punctuation. This feature is ranked 3 out of 180 in later features evaluation.
\end{itemize}

\subsection{All Features Used in Models}
To avoid overfitting problem, we remove all the word n-grams features, resulting of a total of 180 remaining features. This set include previously mentioned features as well as some features examined from previous literatures. All of these features are categorized into three group: post text related (175 features), target content related (2 features) and relation between post text and target content (3 features). 

\section{Empirical Validation}
\subsection{Dataset}
Our data is provided by the Clickbait Challenge 2017~\cite{potthast:2017b}, the topic of which is the detection of clickbait posts in social media. There are 21,997 labeled samples with a distribution of about 25:75 {\em clickbait} and {\em no-clickbait}. Scores of clickbaits are annotated by at least five crowd-sourcing workers. The posts in the training and test sets are judged on a 4-point scale [0, 0.3, 0.66, 1]. A binary ``truthClass" field is also provided in the dataset indicating whether a sample is {\em clickbait} or {\em no-clickbait}. There is no information provided regarding how this binary labels are generated. It is indeed not based on conventional 0.5 threshold on the mean annotation score. Noticeably, the minimum mean judgment score for the {\em clickbait} class is 0.39 (90 instances in total), while the maximum one for {\em no-clickbait} class is 0.59 (4 instances in total). 

Even though the annotators are presented with both the post text and a link to the target article, they are not forced to read the article contents. We also further argue that the judgments for some of the posts are very subjective to the workers' background, especially their topics of interest. Because these control are limited in crowd-sourcing environments, it is understandable that there are much noise in the dataset. Particularly, there are samples that has a same post text and target article but different mean judgment score (e.g. ``{\tt How meditation is transforming American schools}" has 2 instances with 0.6 and 0.8 mean score). Moreover, it is also observed that there is a portion of posts,  annotated as {\em clickbait}, that are totally align with the target content (e.g. ``{\tt How meditation is transforming American schools}", ``{\tt Will it be Heathrow or Gatwick airport that gets the green light for expansion?}").

\subsection{Experimental Setting}
To do feature engineering, we combine all of the labeled data provided by the challenge organizer, and randomly split it with 7:3 ratio as training and testing sets. All the feature engineering is done on the training set. The target variable in regression and classification models is the mean value of clickbait scores and binary clickbait judgment, respectively. 

\textit{All of the subsequent results are from the mean of 10-fold cross validation on the whole dataset.}

\subsection{Results}
Table.\ref{table:allfeatures} presents the results of the 10-fold cross validation on the whole dataset when applying the total 180 features to different prediction models. We use four different models: Linear Regression, Logistics Regression, Random Forest Regression and Random Forest Classifier. All reported F1 score in tables are only for classification on the clickbait class. In case of logistics regression, the F1-score is 0.56 and ROC-AUC is 0.723. Random Forest Regression witnessed the best performance of about 0.036 MSE and 0.819 ACC. This improvement in MST can be credited to the use of mean judgments as the target variable, rather than binary value of either {\em clickbait} or {\em non-clickbait}.

We subsequently separate all 180 features into three groups, namely post-related, target-related and relationship between them, and evaluate each of them in isolation.

\begin{table}
\centering
\begin{tabular}{|c|c|c|c|}
\hline
AUC & MSE & ACC & F1 \\
\hline
\hline
\multicolumn{4}{|c|}{Linear Regression}\\
\hline
0.693 & 0.038 & 0.817 & 0.55\\
\hline 
\multicolumn{4}{|c|}{Logistics Regression}\\
\hline
0.723 & 0.197 & 0.711 & 0.56\\
\hline
\multicolumn{4}{|c|}{Random Forest Regression}\\
\hline
0.69 & 0.036 & 0.819 & 0.55\\
\hline
\multicolumn{4}{|c|}{Random Forest Classifier}\\ 
\hline
0.69 & 0.036 & 0.818 & 0.54\\
\hline
\hline\end{tabular}
\caption{Performance of the clickbait classifiers using different prediction models} 
\label{table:allfeatures}
\end{table}

\begin{table*}
\centering
\begin{tabular}{|c|c|c|c|c|c|c|c|c|}
\hline
 & \multicolumn{4}{|c|}{Linear Regression} & \multicolumn{4}{|c|}{Logistics Regression} \\ 
\hline
GROUPS & AUC & MSE & ACC & F1 & AUC & MSE & ACC & F1\\ 
\hline
Post-related (175 features) & 0.694 & 0.038 & 0.694 & 0.55 & 0.753 & 0.162 & 0.762 & 0.61\\ 
\hline
Target-related (2 features) & 0.5 & 0.062 & 0.749 & 0 & 0.565 & 0.299 & 0.562 & 0.4\\ 
\hline
Relation (3 features) & 0.5 & 0.06 & 0.749 & 0 & 0.56 & 0.278 & 0.593 & 0.38\\ 
\hline
 & \multicolumn{4}{|c|}{Random Forest Regression}  & \multicolumn{4}{|c|}{Random Forest Classifier} \\ 
\hline
GROUPS & AUC & MSE & ACC & F1 & AUC & MSE & ACC & F1\\ 
\hline
Post-related (175 features) & 0.693 & 0.036 & 0.818 & 0.55 & 0.742 & 0.152 & 0.778 & 0.6\\ 
\hline
Target-related (2 features) & 0.537 & 0.067 & 0.734 & 0.21 & 0.571 & 0.229 & 0.665 & 0.39\\ 
\hline
Relation (3 features) & 0.564 & 0.062 & 0.748 & 0.28 & 0.608 & 0.215 & 0.681 & 0.42\\ 
\hline
\hline\end{tabular}
\caption{Performance of the clickbait classifiers using different prediction models with different feature groups.} 
\label{table:group}
\end{table*}

\begin{savenotes}
\begin{table*}
\centering
\begin{tabular}{|c|c|c|c|}
\hline
Rank & Feature & Rank & Feature\\ 
\hline
1 & Number of NNP\footnote{NNP: Proper noun, singular} & 31 & Count POS pattern DT\footnote{DT: Determiner}\\ 
\hline
2 & Readability of target paragraphs (1) & 32 & Number of DT\\ 
\hline
3 & \textbf{Number of tokens} & 33 & POS 2-gram NNP IN\\ 
\hline
4 & Word length of post text & 34 & POS 3-gram IN NNP NNP\\ 
\hline
5 & POS 2-gram NNP NNP & 35 & Number of POS\footnote{POS: Possessive ending}\\ 
\hline
6 & Whether the post start with number & 36 & POS 2-gram IN NN\\ 
\hline
7 & Average length of words in post & 37 & \textbf{Match between keywords and post}\\ 
\hline
8 & Number of IN\footnote{IN: Preposition or subordinating conjunction} & 38 & Number of ','\\ 
\hline
9 & POS 2-gram NNP VBZ\footnote{VBZ: Verb, 3rd person singular present} & 39 & POS 2-gram NNP NNS\footnote{NNS: Noun, plural}\\ 
\hline
10 & POS 2-gram IN NNP & 40 & POS 2-gram IN JJ\footnote{JJ: Adjective}\\ 
\hline
11 & Length of the longest word in post text & 41 & POS 2-gram NNP POS\\ 
\hline
12 & Number of WRB\footnote{WRB: Wh-adverb} & 42 & Number of WDT\footnote{WDT: Wh-determiner}\\ 
\hline
13 & Count POS pattern WRB & 43 & Count POS pattern WDT\\ 
\hline
14 & Number of NN\footnote{NN: Noun, singular or mass} & 44 & POS 2-gram NN NN\\ 
\hline
15 & Count POS pattern NN & 45 & POS 2-gram NN NNP\\ 
\hline
16 & Whether post text start with 5W1H\footnote{5WIH: What, where, when, which, how} & 46 & POS 2-gram NNP VBD\footnote{VBD: Verb, past tense}\\ 
\hline
17 & Whether exist QM\footnote{QM: Question mark} & 47 & \textbf{Similarity between post and target paragraphs}\\ 
\hline
18 & \textbf{Similarity between post and target title} & 48 & Count POS pattern RB\footnote{RB: Adverb}\\ 
\hline
19 & Count POS pattern this/these NN & 49 & Number of RB\\ 
\hline
20 & Count POS pattern PRP\footnote{PRP: Personal pronoun } & 50 & POS 3-gram NNP NNP NNP\\ 
\hline
21 & Number of PRP & 51 & POS 3-gram NNP NNP NN\\ 
\hline
22 & Number of VBZ & 52 & Readability of target paragraphs (2)\\ 
\hline
23 & POS 3-gram NNP NNP VBZ & 53 & Number of RBS\footnote{RBS: Adverb, superlative}\\ 
\hline
24 & POS 2-gram NN IN & 54 & Number of VBN\footnote{VBN: Verb, past participle}\\ 
\hline
25 & POS 3-gram NN IN NNP & 55 & POS 2-gram VBN IN\\ 
\hline
26 & Ratio of stop words in post text & 56 & Whether exist NUMBER NP\footnote{NP: Noun phrase} VB\footnote{VB: Verb, base form}\\ 
\hline
27 & POS 2-gram NNP . & 57 & POS 2-gram JJ NNP\\ 
\hline
28 & POS 2-gram PRP VBP\footnote{VBP: Verb, non-3rd person singular present} & 58 & POS 3-gram NNP NN NN\\ 
\hline
29 & Count POS pattern WP\footnote{WP: Wh-pronoun} & 59 & POS 2-gram DT NN\\ 
\hline
30 & Number of WP & 60 & Whether exist EX\footnote{EX: Existential there}\\ 
\hline
\hline\end{tabular}

\caption{60 most important features} 
\label{table:featurerank}
\end{table*}
\end{savenotes}

\begin{table}
\centering
\begin{tabular}{|c|c|c|c|}
\hline
AUC & MSE & ACC & F1 \\
\hline
\hline
\multicolumn{4}{|c|}{Linear Regression}\\
\hline
0.684 & 0.039 & 0.813 & 0.53 \\
\hline 
\multicolumn{4}{|c|}{Logistics Regression}\\
\hline
 0.745 & 0.171 & 0.75 & 0.6\\
\hline
\multicolumn{4}{|c|}{Random Forest Regression}\\
\hline
0.701 & 0.035 & 0.82 & 0.56\\
\hline
\multicolumn{4}{|c|}{Random Forest Classifier}\\ 
\hline
 0.745 & 0.151 & 0.781 & 0.61\\
\hline
\hline\end{tabular}
\caption{Performance of the clickbait classifiers using different prediction models with the top 60 features} 
\label{table:selectedfeatures}
\end{table}

Table~\ref{table:group} presents the results of the 10-fold cross validation on the three feature groups. In overall, the post-related group has the best performance. Linear Regression undergo better performance under target-related and relation groups, which can be attributed to the outweigh in the number of in no-clickbait instances. The best MSE performance is 0.036 with Random Forest Regression classifier on the post-related group. The F1-scores for Linear Regression classifier are both 0, which can be explained from the unbalanced in data set between the two {\em clickbait} and {\em no-clickbait} class.

\subsection{Feature Selection and Results}
Only the top 60 features are selected for final submission in order to reduce run time and reduce the variance of the model. In particular, we remove word n-grams features. Then the remained features are ranked according to Fisher score, one of the most widely used supervised features selection methods~\cite{duda2012pattern}. Here we show the top 60 important features in Table~\ref{table:featurerank}. 
The footprints are from POS tags used in the Penn Treebank Project.

Table~\ref{table:selectedfeatures} presents the performance of the selected top 60 important features with different models. With features selection, Random Forest Classifier see an improvement of F1-score to 0.61. Similarly, MSE of Random Forest Regression increase to 0.35. 

\subsection{Error Analysis}
In this section, we will examine some of samples that are misclassified by Random Forest Regression model with 400 estimators (trees) and a maximum depth of 20. We first train the model on our training set on the selected top 60 features with the mean judgment as response variable and subsequently analyze the errors on the testing set. We use 0.5 as the binary classification threshold for {\em clickbait} and {\em non-clickbait} labels.

Analysis shows that among the misclassification, about 48\% of them has the mean annotation score ranging from 0.33 to 0.66. Statistically, the interquartile of such score among misclassified instances falls in the range from 0.27 to 0.6. This shows that even though the model can distinguish between extreme {\em clickbait} and {\em non-clickbait} samples, it has limitation in classifying ambiguous instances. This is expected because annotation for these cases are challenging even to human beings. 

\section{Conclusion}
We extract 331 features in total and keep 180 features to avoid overfitting. The top 60 features are selected to reduce run time and decrease noise interference. We apply the 60 features to several machine learning models, which is to generate score of clickbait that rates how click baiting a social media post is. The original six features we contribute are strong indicators according to the rank of feature importance.

Nevertheless, there are still space for improvement. Our future directions include: (1) Extracting more features: word embedding, formality of both post text and target paragraphs, and potential features from associated media; (2) Experimenting other machine learning models, especially deep learning; (3) Collecting more high-quality data: we are planning to expand the existing dataset not only in quantity, but also in vertical dimensions, to capture more information such as URL information or users comments.

\begin{raggedright}
\bibliography{clickbait17-notebook-lit}

\begin{thebibliography}{4}
\providecommand{\natexlab}[1]{#1}
\providecommand{\url}[1]{\texttt{#1}}
\expandafter\ifx\csname urlstyle\endcsname\relax
  \providecommand{\doi}[1]{doi: #1}\else
  \providecommand{\doi}{doi: \begingroup \urlstyle{rm}\Url}\fi

\bibitem[Potthast et~al.(2014)Potthast, Gollub, Rangel, Rosso, Stamatatos, and
  Stein]{potthast:2014}
M.~Potthast, T.~Gollub, F.~Rangel, P.~Rosso, E.~Stamatatos, and B.~Stein.
\newblock {Improving the Reproducibility of PAN's Shared Tasks: Plagiarism
  Detection, Author Identification, and Author Profiling}.
\newblock In \emph{{CLEF}}, pages 268--299. Springer, 2014.

\bibitem[Potthast et~al.(2016)Potthast, K{\"o}psel, Stein, and
  Hagen]{potthast:2016}
M.~Potthast, S.~K{\"o}psel, B.~Stein, and M.~Hagen.
\newblock {Clickbait Detection}.
\newblock In N.~Ferro, F.~Crestani, M.-F. Moens, J.~Mothe, F.~Silvestri, G.~{Di
  Nunzio}, C.~Hauff, and G.~Silvello, editors, \emph{Advances in Information
  Retrieval. 38th European Conference on IR Research (ECIR 16)}, volume 9626 of
  \emph{Lecture Notes in Computer Science}, pages 810--817, Berlin Heidelberg
  New York, Mar. 2016. Springer.
\newblock \doi{http://dx.doi.org/10.1007/978-3-319-30671-1_72}.

\bibitem[Potthast et~al.(2017{\natexlab{a}})Potthast, Gollub, Hagen, and
  Stein]{potthast:2017a}
M.~Potthast, T.~Gollub, M.~Hagen, and B.~Stein.
\newblock {The Clickbait Challenge 2017: Towards a Regression Model for
  Clickbait Strength}.
\newblock In \emph{{Proceddings of the Clickbait Chhallenge}},
  2017{\natexlab{a}}.

\bibitem[Potthast et~al.(2017{\natexlab{b}})Potthast, Gollub, Komlossy,
  Schuster, Wiegmann, Garces, Hagen, and Stein]{potthast:2017b}
M.~Potthast, T.~Gollub, K.~Komlossy, S.~Schuster, M.~Wiegmann, E.~Garces,
  M.~Hagen, and B.~Stein.
\newblock {Crowdsourcing a Large Corpus of Clickbait on Twitter}.
\newblock In \emph{{(to appear)}}, 2017{\natexlab{b}}.

\end{thebibliography}
\end{raggedright}
\end{document}